\documentclass[a4paper, 10pt, twocolumn, twoside]{article}
\usepackage{amssymb}

\usepackage{ISARC}

\usepackage{lscape}
\usepackage{hologo}
\usepackage[none]{hyphenat}

\begin{document}

 % Do not change the following line
\linespread{0.5}
\setlength{\tabcolsep}{4pt} % Default value: 6pt

\title{KUKAloha: A General, Low-Cost, and Shared-Control based Teleoperation Framework for Construction Robot Arm}

\author{Yifan Xu$^1$, Qizhang Shen$^2$, Vineet Kamat$^3$ and Carol Menassa$^4$}

\affiliation{
$^1$Department of Civil and Environmental Engineering, University of Michigan, United States of America\\
$^2$Department of Robotics, University of Michigan, United States of America\\
$^3$Department of Civil and Environmental Engineering, University of Michigan, United States of America\\
$^4$Department of Civil and Environmental Engineering, University of Michigan, United States of America
}

\email{
\href{mailto:e.yfx@umich.edu}{yfx@umich.edu}, 
\href{mailto:e.qzshen@umich.edu}{qzshen@umich.edu},
\href{mailto:e.vkamat@umich.edu}{vkamat@umich.edu},
\href{mailto:e.menassa@umich.edu}{menassa@umich.edu}
}

% Do not change the following three lines
\maketitle 
\thispagestyle{fancy} 
\pagestyle{fancy}

\begin{abstract}
This paper presents KUKAloha, a general, low-cost, and shared-control teleoperation framework designed for construction robot arms. The proposed system employs a leader–follower paradigm in which a lightweight leading arm enables intuitive human guidance for coarse robot motion, while an autonomous perception module based on AprilTag detection performs precise alignment and grasp execution. By explicitly decoupling human control from fine manipulation, KUKAloha improves safety and repeatability when operating large-scale manipulators. We implement the framework on a KUKA robot arm and conduct a usability study with representative construction manipulation tasks. Experimental results demonstrate that KUKAloha reduces operator workload, improves task completion efficiency, and provides a practical solution for scalable demonstration collection and shared human–robot control in construction environments.

\end{abstract}


\section{Introduction}
Construction sites remain among the most hazardous and labor-intensive work environments, with workers routinely exposed to safety risks arising from heavy materials, repetitive operations, constrained spaces, and dynamic site conditions~\cite{Kumar2024}. Despite recent advances in construction automation, many tasks—such as material handling, drilling, fastening, and component assembly—continue to rely heavily on manual labor~\cite{Xu2025}. Robotic manipulators offer a promising pathway to improve safety, productivity, and consistency by offloading physically demanding and high-risk operations from human workers~\cite{Yu2023,Wang2024}. However, the deployment of robot arms in construction environments remains limited due to challenges in usability, adaptability, and data-efficient skill acquisition~\cite{Zeng2024}.

Imitation learning has emerged as a powerful paradigm for enabling robots to acquire complex manipulation skills by learning directly from human demonstrations~\cite{Fang2019SurveyOI,Hussein2017}. Compared to classical programming or purely autonomous planning, imitation learning allows robots to capture nuanced task strategies and adapt to unstructured environments that are common in construction sites. Nevertheless, the effectiveness of imitation learning is fundamentally constrained by the quality, scale, and diversity of demonstration data~\cite{Belkhale2023}. Collecting large amounts of high-quality demonstrations for construction robot arms is particularly challenging due to the size of the robots, the complexity of construction tasks, and the lack of intuitive and accessible teleoperation interfaces.

Human teleoperation has consistently proven to be one of the best ways to control robot manipulators to collect demonstrations for imitation learning~\cite{wu2023gello}. Existing teleoperation solutions for construction robot arms primarily rely on augmented or virtual reality (AR/VR) systems~\cite{Park2025}, teach pendants~\cite{Fukui2009}, or 3D mouse~\cite{Dhat2024}. While these approaches are functional, they often impose a steep learning curve, require specialized hardware, or lack the intuitiveness and precision needed for fine-grained manipulation. Teach pendants and 3D mouse are designed for industrial programming rather than continuous and diverse demonstration. AR/VR systems introduce additional cost and setup complexity and are poorly suited for real-time human-in-the-loop operation and cannot be suitable for diverse demonstrations. As a result, these methods are suboptimal for scalable demonstration collection and shared human–robot control in realistic construction scenarios.

Recent teleoperation systems such as Aloha~\cite{fu2024mobile} and Gello~\cite{wu2023gello} have demonstrated that low-cost, physically grounded, and highly intuitive interfaces can significantly improve the efficiency and naturalness of robot arm control. By enabling direct kinesthetic mapping between human motion and robot motion, these systems lower the barrier to teleoperation and facilitate the collection of high-quality demonstrations for imitation learning~\cite{Zhao2023LearningFB}. However, existing systems are largely developed and evaluated in laboratory settings~\cite{wu2023gello}, focusing on lightweight manipulators and household manipulation tasks. Their direct application to construction robot arms—such as industrial-grade KUKA manipulators—poses additional challenges related to scale, safety, workspace constraints, and the need for shared control to balance human intent with robot autonomy.

To address these challenges, this paper introduces KUKAloha, a general, low-cost, and shared-control-based teleoperation framework designed specifically for construction robot arms. Unlike purely human-driven teleoperation systems, KUKAloha adopts a shared-control paradigm that combines human intuition with autonomous perception and control. In the proposed framework, the human operator provides high-level guidance to maneuver the robot arm toward the target object, leveraging intuitive teleoperation for coarse positioning and situational awareness. Once the robot reaches the vicinity of the object, an autonomous perception module based on AprilTag detection is activated to perform precise pose estimation, enabling accurate final alignment and reliable grasp execution. This hybrid strategy allows KUKAloha to balance flexibility and precision, reducing operator workload while improving task consistency and safety during manipulation.

The main contributions of this paper are summarized as follows:

\begin{itemize}
    \item We propose KUKAloha, a unified leader–follower teleoperation framework that generalizes across different robot arms, end-effectors, and construction tasks.
    \item We introduce a shared-control strategy that combines intuitive human teleoperation for coarse motion with autonomous perception-driven alignment for fine manipulation.
    \item We conduct a comprehensive usability study of KUKAloha in representative construction manipulation tasks, evaluating system usability, operator workload, intuitiveness, and task completion efficiency, and comparing the proposed framework against commonly used teleoperation approaches.
\end{itemize}

\section{Related Work}
\label{sec:relatedwork}
\par 

\begin{figure}
    \centering
    \includegraphics[width=\linewidth]{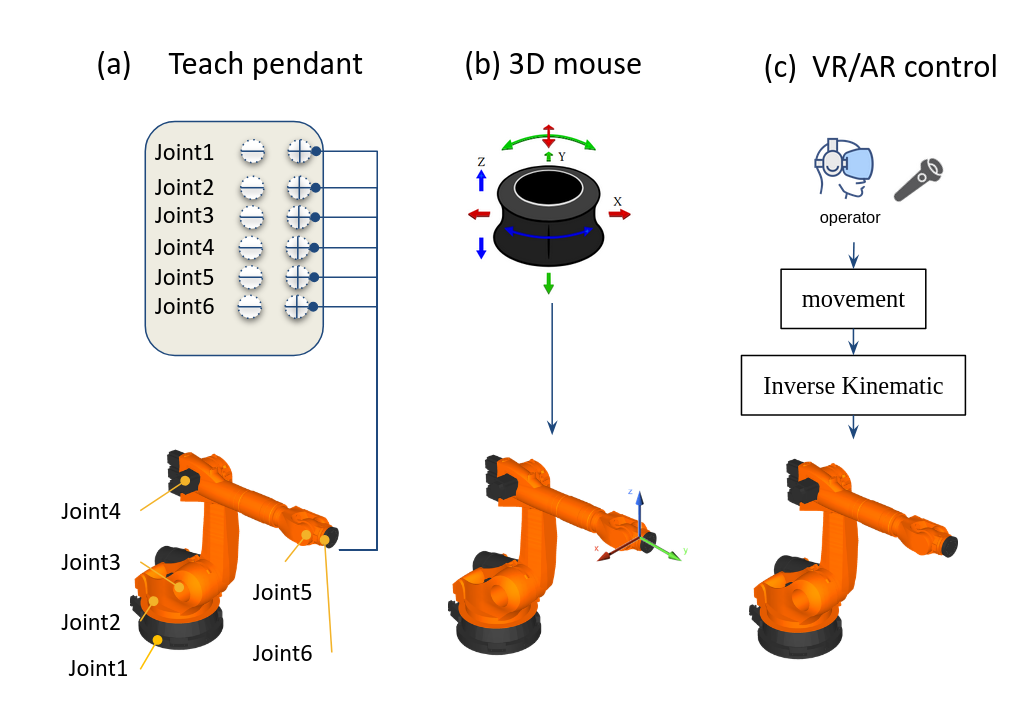}
    \caption{The current ways of teleoperation of construction robot}
    \label{fig:relatedwork}
\end{figure}

Prior work on construction robot arm teleoperation can be broadly categorized into teach pendant–based control~\cite{Fukui2009}, 3D mouse control~\cite{Dhat2024}, and immersive VR/AR-based interfaces~\cite{Park2025}. While each approach has demonstrated effectiveness in specific contexts, significant limitations remain when these methods are applied to construction-scale robot arms and demonstration-driven learning. The demonstration of the control methods is shown in Figure~\ref{fig:relatedwork}.

\subsection{Teach Pendant Teleoperation}

Teach pendants are the most widely used interface for industrial robot programming and manipulation. They provide direct access to low-level robot commands, allowing operators to move the robot incrementally in joint or Cartesian space, record waypoints, and define task sequences~\cite{Fang2019SurveyOI}. Teach pendants are robust, reliable, and well integrated into industrial robot ecosystems, making them a standard tool in manufacturing and construction automation~\cite{Hsieh}. However, teach pendant control is primarily designed for offline programming rather than continuous, intuitive teleoperation. Operation typically requires expert knowledge of robot kinematics and coordinate frames, leading to a steep learning curve for non-expert users~\cite{Ajaykumar2023}. Moreover, the discrete and button-based interaction paradigm limits motion fluidity, making it inefficient for real-time demonstration collection and fine manipulation. 

\subsection{3D Mouse and Motion-Based Input Devices}

3D input devices, such as six-degree-of-freedom (6-DoF) mice and motion controllers, have been explored as an alternative to teach pendants for robot teleoperation~\cite{Dhat2024}. These devices allow operators to specify translational and rotational commands simultaneously, enabling smoother Cartesian control compared to button-based interfaces. Prior studies have shown that 3D mice can improve efficiency and intuitiveness for tasks requiring continuous motion, such as positioning and alignment. Despite these advantages, 3D mouse–based teleoperation remains largely indirect. The lack of physical embodiment and kinesthetic feedback makes it difficult for operators to develop an intuitive sense of robot motion, particularly for large-scale manipulators operating in constrained construction environments~\cite{wu2023gello}. Additionally, precise control near contact remains challenging, often requiring mode switching or careful tuning of control gains. As a result, 3D mouse interfaces offer limited support for high-quality demonstration collection and shared control.

\subsection{VR/AR-Based Teleoperation}

Virtual reality (VR) and augmented reality (AR) interfaces have gained increasing attention for robot teleoperation due to their immersive visualization capabilities~\cite{Park2025}. By providing rich spatial context and natural hand-tracking or controller-based inputs, VR/AR systems can improve operator situational awareness and reduce cognitive load~\cite{Baumeister2017}. Several works have demonstrated the effectiveness of VR-based teleoperation for complex manipulation tasks, including assembly and remote inspection~\cite{Kadavasal2009,Sun2020,Barentine2021}.

However, VR/AR systems introduce additional hardware requirements, calibration complexity, and cost, which can limit their practicality for deployment on construction sites~\cite{DavilaDelgado2020}. Most VR teleoperation systems operate by mapping operator end-effector commands into robot motion through inverse kinematics (IK) solvers that compute the required joint positions to achieve the desired pose of the robot’s end-effector. At singular configurations, the robot loses certain instantaneous motion capabilities, and command mapping can result in unstable or unsafe behavior without additional safeguards~\cite{Ortenzi2019}. Furthermore, VR/AR interfaces often prioritize visualization over physical interaction, making them less suitable for intuitive kinesthetic demonstration and rapid data collection for imitation learning.

\subsection{Research Gap}

In summary, existing teleoperation approaches for robot arms present a trade-off between robustness, intuitiveness, cost, and scalability. Teach pendants are reliable but unintuitive for real-time demonstration; 3D input devices offer smoother control but lack embodiment; and VR/AR systems provide immersion at the expense of complexity and accessibility. These limitations are particularly pronounced for construction robot arms, where large workspace, safety considerations, and the need for scalable demonstration collection pose additional challenges.

In contrast, KUKAloha builds upon physically grounded, leader–follower teleoperation principles to provide an intuitive, low-cost, and shared-control framework tailored for construction environments. By combining human-guided motion with autonomous perception-based precision assistance, the proposed approach addresses key gaps in existing teleoperation systems and supports both practical deployment and learning-based manipulation.

\section{Methodology}
\label{sec:methodology}
\par 

\begin{figure}
    \centering
    \includegraphics[width=1.2\linewidth]{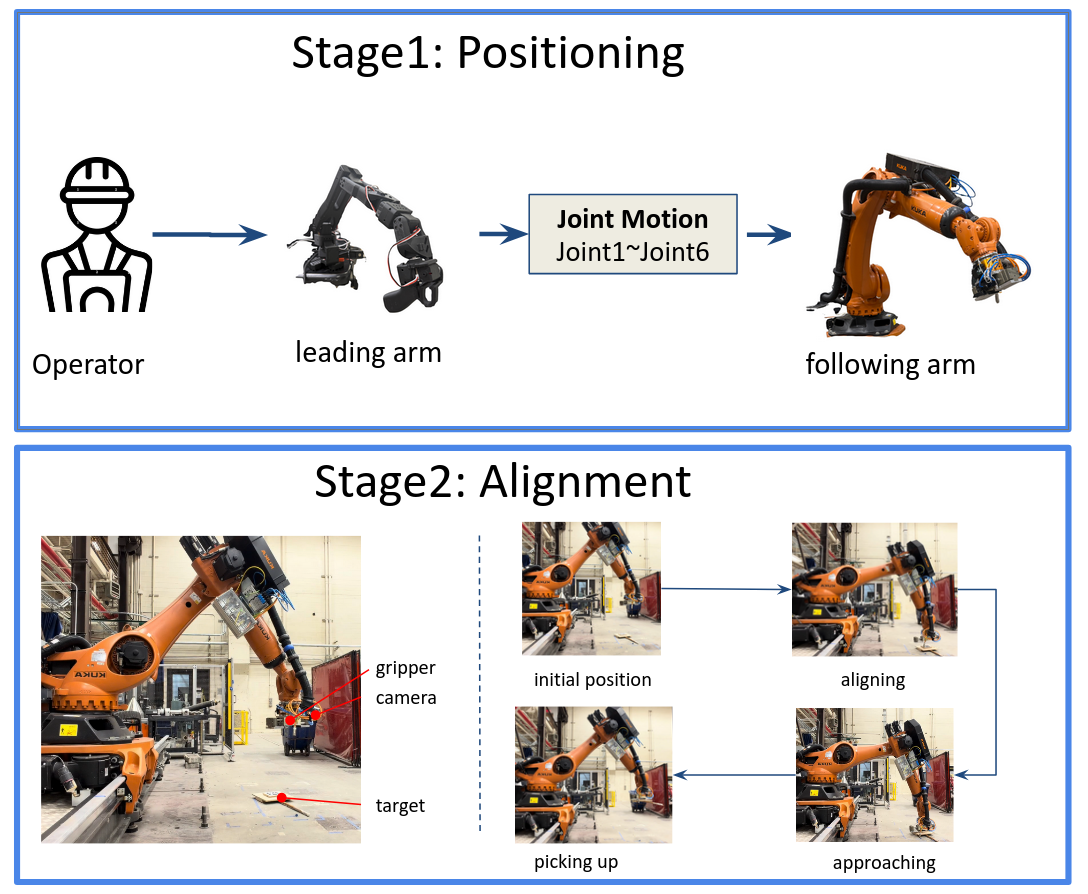}
    \caption{KUKAloha system overview}
    \label{fig:system_overview}
\end{figure}

KUKAloha is designed as a shared-control teleoperation framework that combines intuitive human guidance with autonomous perception-driven precision. The system follows a two-stage control pipeline: (1) a leader–follower teleoperation stage for coarse robot motion, and (2) an autonomous alignment and pickup stage based on visual fiducial detection. The design framework is shown in Figure~\ref{fig:system_overview}. This design leverages the strengths of both human intuition and robot autonomy while mitigating safety and precision limitations inherent in pure teleoperation.

\subsection{Stage I: Leader–Follower Teleoperation Using a Leading Arm}
Stage I provides the human operator with direct, intuitive control over the large-scale robot arm to accomplish coarse positioning tasks in complex construction environments. Rather than requiring precise manual alignment or contact-level control, the objective of this stage is to efficiently maneuver the robot into a task-relevant pre-grasp configuration. By restricting human control to global motion guidance, Stage I reduces operator workload while preserving flexibility and situational awareness.

\begin{figure}[t]
    \centering
    \includegraphics[width=0.55\linewidth]{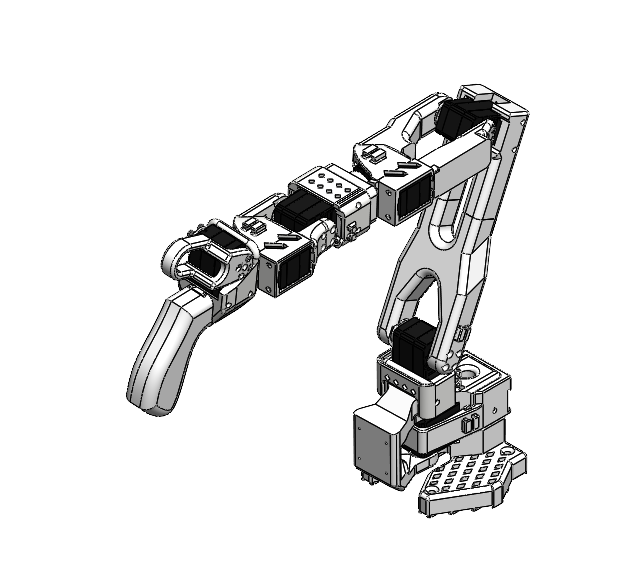}
    \caption{Custom 7-DoF leader arm used as the teleoperation interface. 
    The tabletop device is approximately $1\!:\!7$ scale relative to the KUKA follower manipulator and is operated kinesthetically by the user via the contoured handle at the distal link.}
    \label{fig:leader_arm}
\end{figure}

\paragraph{Mechanical Design of the Leader Arm}
As illustrated in Fig.~\ref{fig:leader_arm}, the leader interface is a compact 7-DoF tabletop arm designed as a physically grounded input device rather than a high-power manipulator. The base module is rigidly mounted on a table and supports two larger proximal links that mimic the shoulder and elbow structure of the KUKA follower arm. These are followed by a chain of wrist joints terminating in a contoured handle that the operator grasps during teleoperation. The kinematic structure and joint ordering mirror those of the KUKA manipulator so that each revolute axis has a direct counterpart, enabling the one-to-one joint-space mapping in \eqref{eq:leader_follower_mapping}. The overall geometry is approximately $1\!:\!7$ scale relative to the follower robot, which allows the leader arm to fit comfortably on a desktop while spanning a workspace that meaningfully represents the follower’s motion. Each joint is driven by a low-power servomotor with integrated position sensing, and the links are built from lightweight structural components to keep inertia and impact forces low. The mechanical layout and software interface draw practical inspiration from the open-source LeRobot SO101 leader designs~\cite{lerobot}, but are adapted to the larger KUKA arm and construction setting.

\paragraph{Joint-Space Mapping and Motion Planning}
In the proposed framework, teleoperation is realized as a joint-space leader--follower mapping between the lightweight leader arm interface and the KUKA follower manipulator. Let $\mathbf{q}_{L}(t) \in \mathbb{R}^{n}$ denote the vector of joint positions measured from the leader arm at time $t$, and let $\mathbf{q}_{B}(t) \in \mathbb{R}^{n}$ denote the joint positions of the follower robot, expressed in the robot base joint space. During teleoperation, the leader encoders are read continuously and used to form a desired joint configuration for the follower robot:
\begin{equation}
    \mathbf{q}_{B}^{\mathrm{des}}(t) = \mathbf{q}_{L}(t).
    \label{eq:leader_follower_mapping}
\end{equation}
That is, each leader joint is mapped one-to-one to the corresponding follower joint without software-based scaling. The leader arm is physically scaled down relative to the KUKA manipulator, so the difference in link dimensions provides inherent motion scaling through hardware design while preserving an intuitive kinesthetic correspondence between human motion and robot motion.

The desired joint vector $\mathbf{q}_{B}^{des}(t)$ is not executed instantaneously, but is instead passed to the motion planning framework, which generates a time-parameterized joint trajectory from the current configuration $\mathbf{q}_{B}(t)$ to the target configuration $\mathbf{q}_{B}^{des}(t)$:
\begin{equation}
    \tau(t) = 
    \text{PlanJointTrajectory}\!\left(
        \mathbf{q}_{B}(t),\,
        \mathbf{q}_{B}^{des}(t)
    \right).
\end{equation}
This trajectory $\tau(t)$ is streamed to a MoveIt-compatible joint trajectory controller, which enforces joint limits, velocity constraints, and collision avoidance while tracking the leader-specified configuration. As a result, the human operates purely in joint space through the leader device, while low-level execution on the construction-scale manipulator remains safe and compliant with industrial control requirements.

Prior to teleoperation, the robot is explicitly switched into the trajectory-based control mode to ensure compatibility with the joint-space teleoperation interface. Teleoperation is intentionally limited to coarse positioning: the operator uses the leader arm to bring the KUKA manipulator into a pre-grasp configuration where the target object and its attached AprilTag are reliably visible to the onboard camera. Once such a configuration is reached, the leader--follower mapping is disengaged and the system transitions to the autonomous alignment and pickup procedure described in Stage~II.

\paragraph{Gravity Compensation on the Leader Arm}
To make the leader arm feel lightweight and comfortable to operate, we compensate for its own gravity and joint friction. Let $\mathbf{q}_{L}$, $\dot{\mathbf{q}}_{L}$, and $\ddot{\mathbf{q}}_{L}$ denote the leader arm joint positions, velocities, and accelerations, respectively. Following the standard rigid-body dynamics formulation for robot manipulators~\cite{lynch2017modern}, the joint-space dynamics of the leader arm can be written as
\begin{equation}
\begin{aligned}
    \boldsymbol{\tau}_{\mathrm{grav}}
    &= \mathbf{M}(\mathbf{q}_{L})\,\ddot{\mathbf{q}}_{L}
     + \mathbf{C}(\mathbf{q}_{L},\dot{\mathbf{q}}_{L})\,\dot{\mathbf{q}}_{L}
     + \mathbf{g}(\mathbf{q}_{L}) \\
    &= \mathrm{RNEA}(\mathbf{q}_{L},\dot{\mathbf{q}}_{L},\ddot{\mathbf{q}}_{L}),
\end{aligned}
\label{eq:grav_comp}
\end{equation}
where $\mathbf{M}(\mathbf{q}_{L})$ is the joint-space inertia matrix, $\mathbf{C}(\mathbf{q}_{L},\dot{\mathbf{q}}_{L})$ is the Coriolis and centrifugal matrix, and $\mathbf{g}(\mathbf{q}_{L})$ is the gravity vector. In practice, we use an inverse-dynamics routine based on the recursive Newton--Euler algorithm (RNEA) to compute $\boldsymbol{\tau}_{\mathrm{grav}}$ for the current joint state of the leader arm.

\paragraph{Friction Compensation}
In addition to gravity, we compensate for joint friction in the leader arm to reduce perceived stick--slip and make small motions easier to perform. For each joint $i$, friction is modeled with separate static and viscous components,
\begin{equation}
    \tau_{\mathrm{fric},i}(\dot{q}_{L,i}) =
    \begin{cases}
        k_{\mathrm{st},i}\,\mathrm{sgn}(\dot{q}_{L,i}), & |\dot{q}_{L,i}| < \dot{q}_{\mathrm{th},i},\\[4pt]
        k_{\mathrm{visc},i}\,\dot{q}_{L,i}, & |\dot{q}_{L,i}| \ge \dot{q}_{\mathrm{th},i},
    \end{cases}
    \label{eq:tau_fric}
\end{equation}
where $k_{\mathrm{st},i}$ is the static friction coefficient, $k_{\mathrm{visc},i}$ is the viscous friction coefficient, and $\dot{q}_{\mathrm{th},i}$ is a small velocity threshold distinguishing the low-speed static-friction regime from the viscous-friction regime. The full friction compensation vector is
\begin{equation}
    \boldsymbol{\tau}_{\mathrm{fric}}(\dot{\mathbf{q}}_{L})
    = \bigl[\tau_{\mathrm{fric},1},\dots,\tau_{\mathrm{fric},n}\bigr]^\top.
\end{equation}
These torques are added to the gravity compensation to reduce resistive forces in the joints and improve the responsiveness of the leader arm during teleoperation.

\paragraph{Joint Difference and Limit Compensation}
To keep the leader arm within a safe and comfortable workspace and to discourage motions that the follower robot cannot reach, we apply a small feedback torque based on the joint difference between the leader and follower. Let $\mathbf{q}_{L}(t)$ and $\mathbf{q}_{B}(t)$ denote the leader and follower joint positions, and define the joint error
\begin{equation}
    \mathbf{e}(t) = \mathbf{q}_{L}(t) - \mathbf{q}_{B}(t).
\end{equation}
When the magnitude of this error exceeds a small threshold and the leader joint is close to its soft limits, a restoring torque
\begin{equation}
    \boldsymbol{\tau}_{\mathrm{joint}}(t) \approx K_{p}\,\mathbf{e}(t)
\end{equation}
is applied on the corresponding leader joints, with additional damping and integral terms in implementation to ensure smooth behavior. This torque gently pulls the leader arm back toward the follower configuration and away from joint limits, providing haptic guidance without restricting free motion in the interior of the workspace.

\paragraph{Trigger-Based Gripper Control}
The leader arm handle includes a trigger input used by the operator to command grasp and release. In our prototype, the trigger is treated as a binary input that issues open/close commands to the follower gripper, decoupled from the joint-space teleoperation of the arm itself. This allows the operator to control object pickup and release through a natural “pull to grasp, release to open” interaction, while the leader arm torques described above handle gravity compensation and joint-space feedback.

\paragraph{Overall Leader Arm Torque Command}
Combining the above components, the torque command applied to the
leader arm servomotors is
\begin{equation}
    \boldsymbol{\tau}_{L}
    = \boldsymbol{\tau}_{\mathrm{grav}}(\mathbf{q}_{L})
    + \boldsymbol{\tau}_{\mathrm{fric}}(\dot{\mathbf{q}}_{L})
    + \boldsymbol{\tau}_{\mathrm{joint}}(\mathbf{q}_{L},\mathbf{q}_{B})
    + \boldsymbol{\tau}_{\mathrm{trig}},
\end{equation}
where $\boldsymbol{\tau}_{\mathrm{trig}}$ is nonzero only for the trigger joint. These terms make the leader arm behave as a low-inertia, physically grounded input device: gravity compensation and friction compensation reduce apparent weight and stiction, joint-limit avoidance provides haptic guidance near workspace boundaries, and the trigger torque is used to render grasp-related force feedback.

Overall, this hardware design allows the leader arm to behave as a low-inertia, physically grounded input device that preserves intuitive joint-space control while supporting the gravity, friction, and limit-compensation torques described above.

\subsection{Stage II: Autonomous Alignment and Pickup via AprilTag Detection}

After coarse positioning via leader--follower teleoperation (Stage I), KUKAloha transitions to an autonomous manipulation stage to achieve precise alignment and safe grasp execution. This stage explicitly decouples fine manipulation from human control to improve repeatability and safety when operating large-scale construction robot arms.

\paragraph{Step 1: Pre-grasp Positioning}
The operator teleoperates the robot to a pre-grasp configuration where the AprilTag attached to the target object is detectable by the onboard camera. 

\paragraph{Step 2: AprilTag-Based Pose Estimation}
Once the AprilTag is detected, the system estimates the 6-DoF pose of the object. Let
\begin{equation}
{}^{C}\mathbf{T}_{tag}
\end{equation}
be the pose of the AprilTag in the camera frame $C$, obtained from the detection pipeline. Given the known camera-to-end-effector extrinsic calibration
\begin{equation}
{}^{E}\mathbf{T}_{C},
\end{equation}
the object pose in the robot base frame is computed as:
\begin{equation}
{}^{B}\mathbf{T}_{obj}
=
{}^{B}\mathbf{T}_{E}
\cdot
{}^{E}\mathbf{T}_{C}
\cdot
{}^{C}\mathbf{T}_{tag}.
\end{equation}

\paragraph{Step 3: Disconnect Leader--Follower Teleoperation}
After a reliable object pose estimate is obtained, the leader--follower mapping is disengaged to remove direct human input from the control loop. The end-effector pose at the moment of disconnection is stored as:
\begin{equation}
{}^{B}\mathbf{T}_{E}^{disc}.
\end{equation}

\paragraph{Step 4: Autonomous Alignment and Grasp Execution}
Using the estimated object pose, the desired grasp pose of the end-effector is defined as:
\begin{equation}
{}^{B}\mathbf{T}_{E}^{grasp}
=
{}^{B}\mathbf{T}_{obj}
\cdot
{}^{obj}\mathbf{T}_{grasp},
\end{equation}
where ${}^{obj}\mathbf{T}_{grasp}$ represents the predefined grasp offset in the object frame.

The robot autonomously executes a Cartesian trajectory toward ${}^{B}\mathbf{T}_{E}^{grasp}$ while respecting joint limits, velocity constraints, and collision avoidance. Upon reaching the target pose, the gripper is actuated to grasp the object.

\paragraph{Step 5: Return to Reconnection Pose}
After a successful grasp, the robot returns to the stored disconnection pose:
\begin{equation}
\mathbf{T}_{E}(t) \rightarrow {}^{B}\mathbf{T}_{E}^{disc},
\end{equation}
ensuring a predictable and safe transition back to human control.

\paragraph{Step 6: Reconnect Leader--Follower Teleoperation}
Finally, the leader--follower teleoperation mapping is re-enabled, allowing the operator to resume intuitive control for subsequent task execution, such as object transport or placement.

\section{Experiment Setup}
\label{sec:result}
\par 

\begin{figure}
    \centering
    \includegraphics[width=\linewidth]{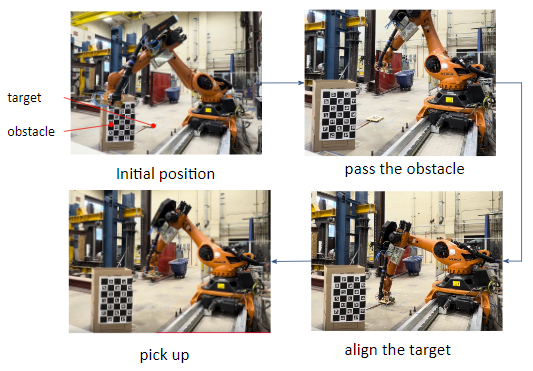}
    \caption{Experiment Setup}
    \label{fig:experiment}
\end{figure}

\subsection{Experiment Scene Setup}

To evaluate the effectiveness of the proposed teleoperation framework, we designed a representative pick-up task using a construction-grade robotic manipulator. The task is intentionally structured to reflect common challenges encountered in construction manipulation scenarios and is decomposed into three sequential stages: (i) obstacle passing, where the robot arm must safely maneuver through a constrained workspace without collision; (ii) target alignment, which requires precise end-effector positioning and orientation relative to the object of interest; and (iii) object pick-up, where stable grasp execution is necessary to successfully lift the target. This staged formulation allows us to assess the teleoperation system’s performance under varying levels of task complexity, from coarse motion planning to fine-grained manipulation. The physical layout of the robot, obstacles, and target object is illustrated in Figure~\ref{fig:experiment}.

The follower robot used in our experiments is a 6-DoF KUKA industrial manipulator equipped with a parallel gripper at the end-effector. For visual perception, a monocular RGB camera is rigidly mounted near the gripper, with its optical axis approximately aligned with the gripping direction. The camera stream is handled by a Raspberry~Pi~4 running a ROS node that forwards image data to the main control computer, where a standard AprilTag detection pipeline estimates the tag pose ${}^{C}\mathbf{T}_{tag}$. The camera-to-end-effector extrinsic transform ${}^{E}\mathbf{T}_{C}$ is obtained via an offline hand--eye calibration procedure and treated as fixed during all experiments.

\subsection{Baseline}

To benchmark the performance of the proposed KUKAloha framework, we compare against several commonly used teleoperation interfaces for construction and industrial robot arms. These baselines include 
\begin{itemize}
    \item VR/AR-based teleoperation
    \item teach pendant control
    \item Leading-Follower arm without autonomous alignment. 
\end{itemize}

which together represent the dominant paradigms currently adopted in industrial and construction robotics.

\subsection{Evaluation Metrics}

To quantitatively evaluate the effectiveness of the proposed teleoperation framework, we measure system performance using several metrics that capture task efficiency, precision, and robustness. These metrics are designed to reflect the requirements of manipulation tasks commonly encountered in construction environments.

\textbf{Task Success Rate.}  
Task success rate measures the percentage of trials in which the robot successfully completes the entire pick-up task without collision or failure. A trial is considered successful if the robot (i) navigates through the obstacle region without contacting any obstacles, (ii) aligns the end-effector with the target object within an acceptable tolerance, and (iii) securely grasps and lifts the object. This metric reflects the overall reliability of the teleoperation system.

\textbf{Task Completion Time.}  
Task completion time measures the elapsed time from the start of the task to successful object pick-up. This metric evaluates the operational efficiency of the teleoperation interface and reflects how quickly a human operator can execute the task using different control methods.

\textbf{Alignment Error.}  
To measure the precision of target alignment, we evaluate both the positional and orientational differences between the desired grasp pose and the executed end-effector pose immediately before grasp execution. Let $\mathbf{p}_{d} = [x_d, y_d, z_d]^T$ and $\mathbf{p}_{ee} = [x_{ee}, y_{ee}, z_{ee}]^T$ denote the desired and actual end-effector positions, respectively. The position alignment error is defined as

\begin{equation}
e_p = \sqrt{(x_{ee}-x_d)^2 + (y_{ee}-y_d)^2 + (z_{ee}-z_d)^2}.
\end{equation}

Let $R_d$ and $R_{ee}$ denote the desired and executed end-effector rotation matrices. The orientation alignment error is computed as

\begin{equation}
e_R = \cos^{-1}\left(\frac{\text{trace}(R_d^T R_{ee}) - 1}{2}\right).
\end{equation}

Lower values of $e_p$ and $e_R$ indicate more accurate end-effector alignment during the fine manipulation stage.

\textbf{Collision Rate.}  
Collision rate measures the frequency of unintended contacts between the robot and the surrounding obstacles during the obstacle passing stage. This metric reflects the safety and controllability of the teleoperation interface.

Together, these metrics allow us to systematically compare the proposed teleoperation framework against the baseline interfaces in terms of efficiency, accuracy, and robustness in representative construction manipulation tasks.

\subsection{Experimental Result}

The quantitative results of the comparison between different teleoperation interfaces are summarized in Table~\ref{tab:result}. We evaluate four methods: VR/AR-based teleoperation, teach pendant control, pure leader–follower teleoperation, and the proposed leader–follower with automatic alignment framework.

\begin{table*}[t]
\centering
\caption{Performance comparison of different teleoperation methods}
\label{tab:result}
\renewcommand{\arraystretch}{3.0}
\begin{tabular}{lccccc}
\hline
Method & Success Rate & Time (s) & Position Error(m) & Orientation Error(rad) & Collision Rate \\
\hline
VR/AR Control & 35\% & 89.57 & 0.33 & 0.704 & 55\% \\
Teach Pendant & \textbf{85\%} & 258.74 & 0.04 & 0.106 & \textbf{0\%} \\
Leader–Follower & 60\% & 77.59 & 0.26 & 0.522 & 10\% \\
KUKAloha (\textbf{Ours}) & 80\% & \textbf{43.56} & \textbf{0.02} & \textbf{0.087} & 5\% \\
\hline
\end{tabular}
\end{table*}

From the results, VR/AR-based teleoperation exhibits the lowest task success rate (35\%) and the highest collision rate (55\%). Although the immersive interface provides intuitive spatial awareness, it introduces latency and control instability that makes precise manipulation difficult in constrained environments. Consequently, the alignment errors are also the largest among all methods ($e_p=0.33$m and $e_R=0.704$ rad).

Teach pendant control achieves the highest task success rate (85\%) and zero collisions due to its precise incremental control. However, it requires significantly longer task completion time (258.74 s), which highlights the inefficiency of manual programming interfaces for real-time manipulation and demonstration collection.

Pure leader–follower teleoperation improves task efficiency compared with the teach pendant and VR/AR control, reducing task completion time to 77.59 s while maintaining moderate success rates (60\%). Nevertheless, without perception assistance, the operator must manually perform fine alignment, resulting in relatively large alignment errors ($e_p=0.26$m and $e_R=0.522$ rad).

The proposed leader–follower with automatic alignment framework achieves the best overall performance. It significantly reduces the task completion time to 43.56 s while maintaining a high success rate (80\%). In addition, it achieves the smallest alignment errors ($e_p=0.02$m and $e_R=0.087$ rad) and a low collision rate (5\%). These results demonstrate that integrating perception-based automatic alignment with intuitive leader–follower teleoperation effectively balances human intuition and robotic precision, leading to faster, safer, and more reliable manipulation.
\section{Conclusion}

This paper presented KUKAloha, a low-cost and shared-control teleoperation framework designed to enable intuitive and efficient operation of construction robot arms. By combining human teleoperation for coarse motion with perception-assisted autonomous alignment, the proposed system bridges the gap between manual robot operation and fully autonomous manipulation. Experimental evaluation on representative pick-up tasks demonstrates that KUKAloha enables reliable and precise manipulation while maintaining intuitive human control. Compared with commonly used teleoperation interfaces such as teach pendants, VR/AR systems, and 3D mouse control, the proposed framework provides a more accessible and practical solution for operating large-scale construction manipulators.

Beyond improving teleoperation usability, KUKAloha also provides an effective platform for collecting high-quality demonstrations to support imitation learning for construction robots. By lowering the barrier to intuitive robot control, the framework enables scalable data collection that can accelerate the development of learning-based manipulation systems for construction tasks. Future work will extend the system to more complex construction operations and integrate learning-based control methods to progressively increase robot autonomy while maintaining effective human–robot collaboration in dynamic construction environments.

\section{Acknowledgement}
\label{sec:acknowledgement}

The work presented in this paper was supported financially by the United States National Science Foundation (NSF) SCC-IRG 2124857. The support of the NSF is gratefully acknowledged.

\bibliography{ISARC}

@article{Kumar2024,
  title = {Investigation of Unsafe Construction Site Conditions Using Deep Learning Algorithms Using Unmanned Aerial Vehicles},
  volume = {24},
  ISSN = {1424-8220},
  url = {http://dx.doi.org/10.3390/s24206737},
  DOI = {10.3390/s24206737},
  number = {20},
  journal = {Sensors},
  publisher = {MDPI AG},
  author = {Kumar,  Sourav and Poyyamozhi,  Mukilan and Murugesan,  Balasubramanian and Rajamanickam,  Narayanamoorthi and Alroobaea,  Roobaea and Nureldeen,  Waleed},
  year = {2024},
  month = oct,
  pages = {6737}
}

@article{Xu2025,
  title = {Automation in manufacturing and assembly of industrialised construction},
  volume = {170},
  ISSN = {0926-5805},
  url = {http://dx.doi.org/10.1016/j.autcon.2024.105945},
  DOI = {10.1016/j.autcon.2024.105945},
  journal = {Automation in Construction},
  publisher = {Elsevier BV},
  author = {Xu,  Li and Zou,  Yang and Lu,  Yuqian and Chang-Richards,  Alice},
  year = {2025},
  month = feb,
  pages = {105945}
}

@article{Yu2023,
  title = {Mutual physical state-aware object handover in full-contact collaborative human-robot construction work},
  volume = {150},
  ISSN = {0926-5805},
  url = {http://dx.doi.org/10.1016/j.autcon.2023.104829},
  DOI = {10.1016/j.autcon.2023.104829},
  journal = {Automation in Construction},
  publisher = {Elsevier BV},
  author = {Yu,  Hongrui and Kamat,  Vineet R. and Menassa,  Carol C. and McGee,  Wes and Guo,  Yijie and Lee,  Honglak},
  year = {2023},
  month = jun,
  pages = {104829}
}

@article{Wang2024,
  title = {Enabling Building Information Model-driven human-robot collaborative construction workflows with closed-loop digital twins},
  volume = {161},
  ISSN = {0166-3615},
  url = {http://dx.doi.org/10.1016/j.compind.2024.104112},
  DOI = {10.1016/j.compind.2024.104112},
  journal = {Computers in Industry},
  publisher = {Elsevier BV},
  author = {Wang,  Xi and Yu,  Hongrui and McGee,  Wes and Menassa,  Carol C. and Kamat,  Vineet R.},
  year = {2024},
  month = oct,
  pages = {104112}
}

@article{Zeng2024,
  title = {Autonomous mobile construction robots in built environment: A comprehensive review},
  volume = {19},
  ISSN = {2666-1659},
  url = {http://dx.doi.org/10.1016/j.dibe.2024.100484},
  DOI = {10.1016/j.dibe.2024.100484},
  journal = {Developments in the Built Environment},
  publisher = {Elsevier BV},
  author = {Zeng,  Lingdong and Guo,  Shuai and Wu,  Jing and Markert,  Bernd},
  year = {2024},
  month = oct,
  pages = {100484}
}

@article{Fang2019SurveyOI,
  title={Survey of imitation learning for robotic manipulation},
  author={Bin Fang and Shi-Dong Jia and Di Guo and Muhua Xu and Shuhuan Wen and Fuchun Sun},
  journal={International Journal of Intelligent Robotics and Applications},
  year={2019},
  volume={3},
  pages={362 - 369},
  url={https://api.semanticscholar.org/CorpusID:202733441}
}

@article{Hussein2017,
  title = {Imitation Learning: A Survey of Learning Methods},
  volume = {50},
  ISSN = {1557-7341},
  url = {http://dx.doi.org/10.1145/3054912},
  DOI = {10.1145/3054912},
  number = {2},
  journal = {ACM Computing Surveys},
  publisher = {Association for Computing Machinery (ACM)},
  author = {Hussein,  Ahmed and Gaber,  Mohamed Medhat and Elyan,  Eyad and Jayne,  Chrisina},
  year = {2017},
  month = apr,
  pages = {1–35}
}

@inproceedings{Belkhale2023,
author = {Belkhale, Suneel and Cui, Yuchen and Sadigh, Dorsa},
title = {Data quality in imitation learning},
year = {2023},
publisher = {Curran Associates Inc.},
address = {Red Hook, NY, USA},
booktitle = {Proceedings of the 37th International Conference on Neural Information Processing Systems},
articleno = {3524},
numpages = {21},
location = {New Orleans, LA, USA},
series = {NIPS '23}
}

@misc{wu2023gello,
    title={GELLO: A General, Low-Cost, and Intuitive Teleoperation Framework for Robot Manipulators},
    author={Philipp Wu and Yide Shentu and Zhongke Yi and Xingyu Lin and Pieter Abbeel},
    year={2023},
}

@article{Park2025,
  title = {Integrating Large Language Models with Multimodal Virtual Reality Interfaces to Support Collaborative Human–Robot Construction Work},
  volume = {39},
  ISSN = {1943-5487},
  url = {http://dx.doi.org/10.1061/JCCEE5.CPENG-6106},
  DOI = {10.1061/jccee5.cpeng-6106},
  number = {1},
  journal = {Journal of Computing in Civil Engineering},
  publisher = {American Society of Civil Engineers (ASCE)},
  author = {Park,  Somin and Menassa,  Carol C. and Kamat,  Vineet R.},
  year = {2025},
  month = jan 
}

@inproceedings{Fukui2009,
  title = {Development of teaching pendant optimized for robot application},
  url = {http://dx.doi.org/10.1109/ARSO.2009.5587070},
  DOI = {10.1109/arso.2009.5587070},
  booktitle = {2009 IEEE Workshop on Advanced Robotics and its Social Impacts},
  publisher = {IEEE},
  author = {Fukui,  Hidetoshi and Yonejima,  Satoshi and Yamano,  Masatake and Dohi,  Masao and Yamada,  Mariko and Nishiki,  Tomonori},
  year = {2009},
  month = nov,
  pages = {72–77}
}

@inproceedings{fu2024mobile,
  author    = {Fu, Zipeng and Zhao, Tony Z. and Finn, Chelsea},
  title     = {Mobile ALOHA: Learning Bimanual Mobile Manipulation with Low-Cost Whole-Body Teleoperation},
  booktitle = {{Conference on Robot Learning (CoRL)}},
  year      = {2024},
}

@inproceedings{Dhat2024,
author = {Dhat, Varad and Walker, Nick and Cakmak, Maya},
title = {Using 3D Mice to Control Robot Manipulators},
year = {2024},
isbn = {9798400703225},
publisher = {Association for Computing Machinery},
address = {New York, NY, USA},
url = {https://doi.org/10.1145/3610977.3637486},
doi = {10.1145/3610977.3637486},
booktitle = {Proceedings of the 2024 ACM/IEEE International Conference on Human-Robot Interaction},
pages = {896–900},
numpages = {5},
location = {Boulder, CO, USA},
series = {HRI '24}
}

@article{Zhao2023LearningFB,
  title={Learning Fine-Grained Bimanual Manipulation with Low-Cost Hardware},
  author={Tony Zhao and Vikash Kumar and Sergey Levine and Chelsea Finn},
  journal={ArXiv},
  year={2023},
  volume={abs/2304.13705},
  url={https://api.semanticscholar.org/CorpusID:258331658}
}

@inproceedings{Hsieh,
  title = {Development of Remote Virtual Teach Pendant for Robot Programming: Lessons Learned},
  url = {http://dx.doi.org/10.18260/1-2--32660},
  DOI = {10.18260/1-2--32660},
  booktitle = {2019 ASEE Annual Conference amp; Exposition  Proceedings},
  publisher = {ASEE Conferences},
  year={2019},
  author = {Hsieh,  Sheng-Jen}
}

@article{Ajaykumar2023,
  title = {Curricula for teaching end-users to kinesthetically program collaborative robots},
  volume = {18},
  ISSN = {1932-6203},
  url = {http://dx.doi.org/10.1371/journal.pone.0294786},
  DOI = {10.1371/journal.pone.0294786},
  number = {12},
  journal = {PLOS ONE},
  publisher = {Public Library of Science (PLoS)},
  author = {Ajaykumar,  Gopika and Hager,  Gregory D. and Huang,  Chien-Ming},
  editor = {Adebisi,  John},
  year = {2023},
  month = dec,
  pages = {e0294786}
}

@article{Baumeister2017,
  title = {Cognitive Cost of Using Augmented Reality Displays},
  volume = {23},
  ISSN = {1077-2626},
  url = {http://dx.doi.org/10.1109/TVCG.2017.2735098},
  DOI = {10.1109/tvcg.2017.2735098},
  number = {11},
  journal = {IEEE Transactions on Visualization and Computer Graphics},
  publisher = {Institute of Electrical and Electronics Engineers (IEEE)},
  author = {Baumeister,  James and Ssin,  Seung Youb and ElSayed,  Neven A. M. and Dorrian,  Jillian and Webb,  David P. and Walsh,  James A. and Simon,  Timothy M. and Irlitti,  Andrew and Smith,  Ross T. and Kohler,  Mark and Thomas,  Bruce H.},
  year = {2017},
  month = nov,
  pages = {2378–2388}
}

@article{Kadavasal2009,
  title = {Sensor Augmented Virtual Reality Based Teleoperation Using Mixed Autonomy},
  volume = {9},
  ISSN = {1944-7078},
  url = {http://dx.doi.org/10.1115/1.3086030},
  DOI = {10.1115/1.3086030},
  number = {1},
  journal = {Journal of Computing and Information Science in Engineering},
  publisher = {ASME International},
  author = {Kadavasal,  Muthukkumar S. and Oliver,  James H.},
  year = {2009},
  month = mar 
}

@article{Sun2020,
  title = {A New Mixed-Reality-Based Teleoperation System for Telepresence and Maneuverability Enhancement},
  volume = {50},
  ISSN = {2168-2305},
  url = {http://dx.doi.org/10.1109/THMS.2019.2960676},
  DOI = {10.1109/thms.2019.2960676},
  number = {1},
  journal = {IEEE Transactions on Human-Machine Systems},
  publisher = {Institute of Electrical and Electronics Engineers (IEEE)},
  author = {Sun,  Da and Kiselev,  Andrey and Liao,  Qianfang and Stoyanov,  Todor and Loutfi,  Amy},
  year = {2020},
  month = feb,
  pages = {55–67}
}

@inproceedings{Barentine2021,
  series = {HRI ’21},
  title = {A VR Teleoperation Suite with Manipulation Assist},
  url = {http://dx.doi.org/10.1145/3434074.3447210},
  DOI = {10.1145/3434074.3447210},
  booktitle = {Companion of the 2021 ACM/IEEE International Conference on Human-Robot Interaction},
  publisher = {ACM},
  author = {Barentine,  Christian and McNay,  Andrew and Pfaffenbichler,  Ryan and Smith,  Addyson and Rosen,  Eric and Phillips,  Elizabeth},
  year = {2021},
  month = mar,
  pages = {442–446},
  collection = {HRI ’21}
}

@article{DavilaDelgado2020,
  title = {Augmented and Virtual Reality in Construction: Drivers and Limitations for Industry Adoption},
  volume = {146},
  ISSN = {1943-7862},
  url = {http://dx.doi.org/10.1061/(ASCE)CO.1943-7862.0001844},
  DOI = {10.1061/(asce)co.1943-7862.0001844},
  number = {7},
  journal = {Journal of Construction Engineering and Management},
  publisher = {American Society of Civil Engineers (ASCE)},
  author = {Davila Delgado,  Juan Manuel and Oyedele,  Lukumon and Beach,  Thomas and Demian,  Peter},
  year = {2020},
  month = jul 
}

@inproceedings{Ortenzi2019,
  title = {Singularity-Robust Inverse Kinematics Solver for Tele-manipulation},
  url = {http://dx.doi.org/10.1109/COASE.2019.8842871},
  DOI = {10.1109/coase.2019.8842871},
  booktitle = {2019 IEEE 15th International Conference on Automation Science and Engineering (CASE)},
  publisher = {IEEE},
  author = {Ortenzi,  Valerio and Marturi,  Naresh and Rajasekaran,  Vijaykumar and Adjigble,  Maxime and Stolkin,  Rustam},
  year = {2019},
  month = aug,
  pages = {1821–1828}
}

@misc{lerobot,
  author       = {Hugging Face},
  title        = {LeRobot: An Open-Source Teleoperation and Robot Learning Framework},
  howpublished = {\url{https://github.com/huggingface/lerobot}},
  year         = {2024},
  note         = {Accessed: 2026-01-12}
}

@book{lynch2017modern,
  author    = {Kevin M. Lynch and Frank C. Park},
  title     = {Modern Robotics: Mechanics, Planning, and Control},
  publisher = {Cambridge University Press},
  address   = {New York, NY, USA},
  year      = {2017},
  edition   = {1}
}

\end{document}